\def\year{2020}\relax
\documentclass[letterpaper]{article} 
\usepackage{aaai20}  
\usepackage{times}  
\usepackage{helvet} 
\usepackage{courier}  
\usepackage[hyphens]{url}  
\usepackage{graphicx} 
\usepackage{graphicx}  
\usepackage{multirow}
\usepackage{graphicx}
\usepackage{amsmath}
\usepackage{amsfonts}

\usepackage{epsfig}
\usepackage{breqn}
\usepackage{algorithm}
\usepackage{algorithmic}
\usepackage{capt-of}
\usepackage{pdfpages}
\usepackage{enumerate}
\usepackage{footnote}
\usepackage{multirow,xcolor}
\usepackage{makecell}

\title{Temporal Interlacing Network}
\author{Hao Shao\textsuperscript{\rm 1,2}\ \ \ \ \ \ Shengju Qian\textsuperscript{\rm 3} \ \ \ \ \ \  Yu Liu\textsuperscript{\rm 3}\thanks{Corresponding author.} \\  
\\ 
\textsuperscript{\rm 1}Tsinghua University \ \ \ \ 
\textsuperscript{\rm 2}SenseTime X-Lab \ \ \ \ 
\textsuperscript{\rm 3}The Chinese University of Hong Kong\\
\texttt{shaoh19@mails.tsinghua.edu.cn,
sjqian@cse.cuhk.edu.hk, yuliu@ee.cuhk.edu.hk}
}
 
\begin{document}

\maketitle

\begin{abstract}
For a long time, the vision community tries to learn the spatio-temporal representation by combining convolutional neural network together with various temporal models, such as the families of Markov chain, optical flow, RNN and temporal convolution. However, these pipelines consume enormous computing resources due to the alternately learning process for spatial and temporal information. One natural question is whether we can embed the temporal information into the spatial one so the information in the two domains can be jointly learned once-only. In this work, we answer this question by presenting a simple yet powerful operator -- temporal interlacing network (TIN). Instead of learning the temporal features, TIN fuses the two kinds of information by interlacing spatial representations from the past to the future, and vice versa.  A differentiable interlacing target can be learned to control the interlacing process. In this way, a heavy temporal model is replaced by a simple interlacing operator. We theoretically prove that with a learnable interlacing target, TIN performs equivalently to the regularized temporal convolution network (r-TCN), but gains 4\% more accuracy with 6x less latency on 6 challenging benchmarks.
These results push the state-of-the-art performances of video understanding by a considerable margin.
Not surprising, the ensemble model of the proposed TIN won the $1^{st}$ place in the ICCV19 - Multi Moments in Time challenge. Code is made available to facilitate further research.\footnote{https://github.com/deepcs233/TIN}
\end{abstract}

\section{Introduction}
With the explosive growth of video data and the increasing applications, the requirements for speed and accuracy of video understanding are gradually growing. It is also one of the major research hot-spots in computer vision in recent years. Visual behavior extends from 2D image space to 3D space-time, which dramatically increases the complexity of behavioral expression and downstream recognition tasks. How to embed the temporal dynamics into spatial representations remains challenging. 

\begin{figure}[t!]
\centering
\includegraphics[width=1.0\linewidth]{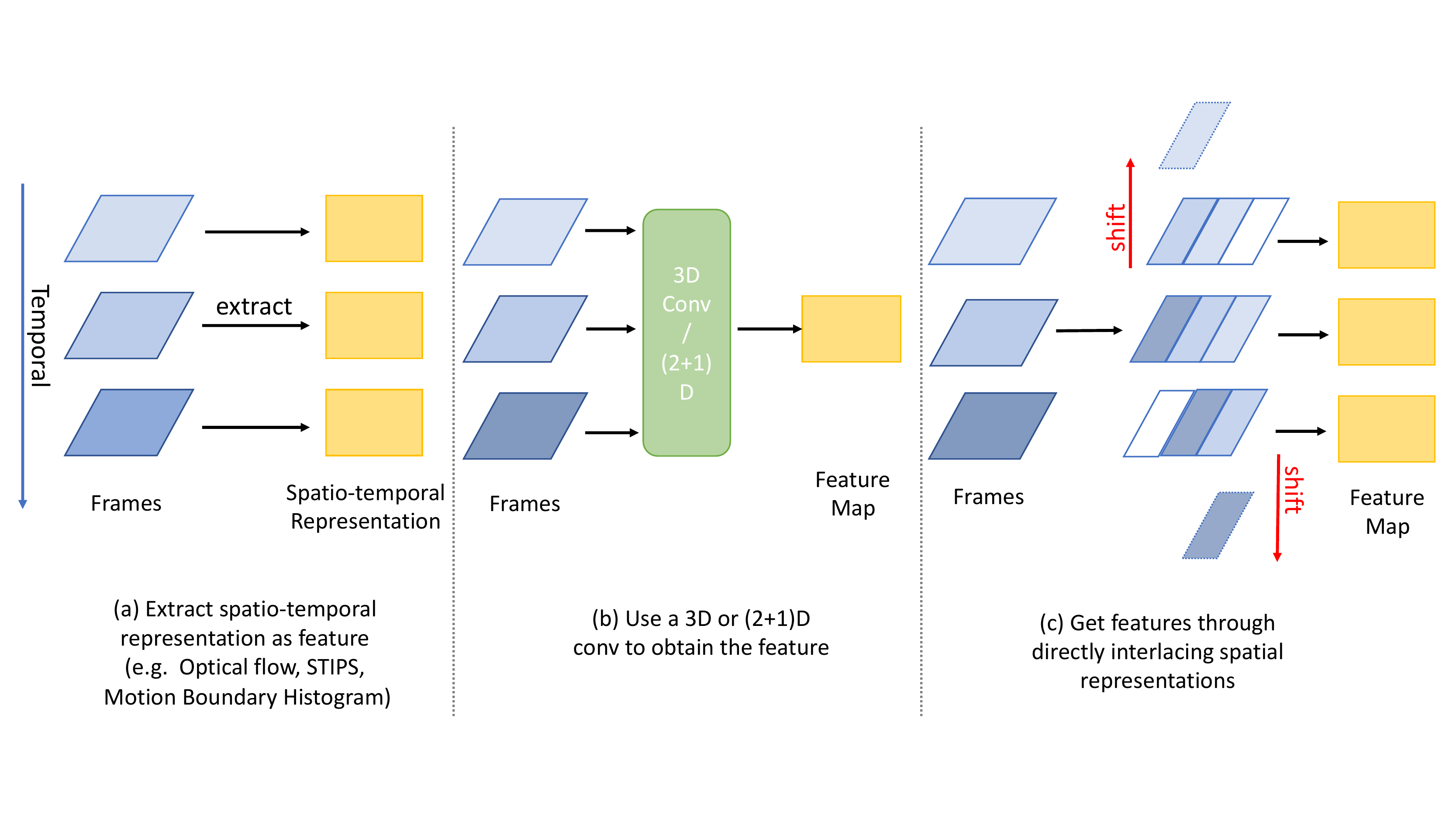}
\caption{\label{pipeline}A brief introduction to three video understanding pipelines. (a) Use hand-crafted features extracted from each frame to help classification. (b) Take a 3D convolution or (2+1)D convolution layer as the temporal feature extractor. (c) Interlace spatial representations of raw frames at different times.} 
\label{fig:pipeline}
\end{figure}

Over the past few decades, researchers try to address video understanding via three pipelines. Before the early deep learning era, hand-crafted temporal representation, e.g. optical flow ~\cite{wang2013action} is commonly used to help the frame representation in video classification, as shown in Figure.~\ref{fig:pipeline}(a). The results are unsatisfying when used alone. After entering the era of deep learning, the second way, as depicted in Figure.~\ref{fig:pipeline}(b), takes neural networks as the feature extractor. The representative methods include LSTM~\cite{ullah2017action}~\cite{donahue2015long}, and network with temporal modeling: C3D~\cite{tran2015learning}, I3D~\cite{carreira2017quo}, R (2+1D)~\cite{tran2018closer}, S3D~\cite{xie2018rethinking}, TSM~\cite{lin2018temporal}, SlowFast~\cite{feichtenhofer2018slowfast}. The convolution kernel is capable of extracting high-dimensional feature and has obtained promising results in many complex benchmarks. The major problems of these solutions change to their vast number of parameters and FLOPs, which make the network hard to converge and prone to overfitting. The third pipeline is proposed by us, as shown in Figure .~\ref{fig:pipeline}(c). It fuses the temporal information via interlacing spatial representations in the temporal dimension. The fused representation can be processed directly in later network layers. This pipeline introduces little extra parameters and FLOPs, maintaining complex interactions in the temporal domain. 

In this paper, we introduce an efficient architecture unit called TIN, which aims at promoting the quality of fusion and modeling with temporal information. The proposed TIN is proved to be equivalent to the regularized temporal convolution network (r-TCN) theoretically. Similarly, TIN is pluggable into any location of the network with little FLOPs and parameters as it retains the spatial size of the input feature map and is extremely lightweight. As shown in Figure. ~\ref{fig:pipeline}(c), TIN directly utilizes interlaced spatial representations instead of extracting features from a heavy network along with temporal dimension. Our architecture is composed of 3 steps: Firstly, our module splits the input channel-wise feature into several groups, obtaining the offsets and weights of neighboring frames to mingle the temporal information. Secondly, we apply the learned offsets to their respective groups through shifting operation and also interpolate the shifted feature along with temporal dimension. Finally, we concatenate the split features and temporal-wisely aggregate them with the learned weights. The grouped offsets can be viewed as separate convolution kernels with different sizes. This design makes our TIN module capable of capturing the long-range temporal relationship and adaptive to various sampling rates across datasets. Besides, the group-wise offsets try to fuse more temporal information at different time stamps. \\

To sum up, we propose an efficient and accurate framework for video recognition with TIN, which has a strong capability of temporal modeling. We also theoretically prove that Temporal Interlacing Network is equivalent to the regularized temporal convolution network (r-TCN). Exhaustive experiments further demonstrate the proposed TIN gains 4\% more accuracy with 6x less latency, and finally be the new state-of-the-art method. Especially, TIN performs as the core architecture in the $1^{st}$ solution of ICCV19 - Multi Moments in Time challenge.

\section{Related Work}
\noindent \textbf{Video Understanding}: With the flourishing development of computing resources and data collecting in recent years, video understanding has caught on dramatically as it provides a wide range of applications. Although CNN architectures have achieved great success on static images, the additional temporal dimension provided in videos inflates the data complexity, the FLOPs of network parameters, and the brittle training procedure significantly. Therefore, how to deal with the information on the temporal dimension, which delivers another important motion cue and alleviates the complexity, is crucial. \\
\noindent Before the beginning of the deep learning era, many studies used some hand-crafted features to help video classification, including HOG3D~\cite{tran2015learning}, SIFT-3D~\cite{scovanner20073}, Action-Bank~\cite{sadanand2012action}. Improved Dense Trajectories (iDT)~\cite{wang2013action} is widely exploited in the field of video understanding, as it provides an efficient video representation and boosts the performance of the models that only use RGB as training data. Two-stream~\cite{simonyan2014two} method obtained a huge improvement with the optical flow input. Since CNN works well on 2D static image understanding, some propose to use 2D CNN directly in video understanding with extended temporal dimensional.TSN proposed to classify the video by fusing the class scores from the frames of different segments. However, it can only fuse the temporal information after computing the class score. The model performs well on datasets that do not depend on temporal relationships (such as K600, UCF101, HMDB51), while it performs poorly on datasets that rely heavily on that (such as something2something v1, something2something v2, jester). The reason is that utilizing 2D CNN for video recognition is essentially exploring static representation learning and still an image classification. It can only classify partial frames in the video and integrate the results. Therefore it is restrained from capturing compact temporal information. For example, in the something2something v1 dataset, the class "Pulling something from left to right" and the class "Pulling something from right to left" are completely inseparable in basic convolution neural network.

\noindent \textbf{Deformable Temporal Modeling}:
Recent works about deformable convolution focus on how to deal with spatial transformations effectively, DCNv1~\cite{dai2017deformable} and DCNv2~\cite{zhu2019deformable} significantly improve performance on the previous state-of-the-arts on semantic segmentation and object detection. Spatial Transformer Networks~\cite{jaderberg2015spatial} use global affine transformation to learn translation-invariant and rotation-invariant feature representation. TSM proposes to shift the feature map along the temporal dimension based on TSN. Nevertheless, it can only shift with fixed displacement and cannot adapt to videos with an uneven number of frames. Inspired by the deformable convolution, we propose a changeable shift operation that can adapt to specific datasets and the distribution of extracted frames.

\noindent \textbf{Self-attention}: We also re-visit the self-attention~\cite{vaswani2017attention} mechanism in this framework. The attention mechanism is originally proposed in the field of machine translation and other natural language processing tasks. The module can be understood as a way of calculating the context in one position using a weighted sum of all positions in a sentence. ~\cite{hu2018squeeze} tried to improve the performance of image classification by modeling the inter-dependencies between the features along the channel dimension. ~\cite{wang2018non} further proposed the non-local neural network for vision tasks such as video classification, object detection, and instance segmentation based on the self-attention method attention to capture long-range dependencies.

\section{Temporal Interlacing Network~(TIN)}

In this section, we introduce the framework of our temporal interlacing network. It is a simple yet effective design that made of two components: deformable shift module and differentiable temporal-wise frame sampling. In the following, we first describe the intuition behind TIN and then explain how to tackle existing problems about temporal modeling in our framework.

\subsection{Intuition}

In temporal datasets, videos often focus on objects or humans performing the interactions. For better capturing the inherent temporal semantics, the critical concern in video understanding compared to image recognition is to model the ``frame-wise'' information. Existing efficient 2D CNN approaches are to relieve the heavy computational cost. However, the trade-offs between learning spatial and temporal features jointly and its heavy computation overhead always exist.

 Previous works focus on extending 2D CNN representation with an extra temporal dimension. In order to mingle features at different times, the temporal convolution network is introduced.  This operation obtains temporal information by looking along the temporal dimension with a fixed visual range. However, a static temporal receptive field removes the computation cost as well as multi-level spatial information. A dynamic temporal receptive field is crucial for jointly embedding the temporal information flow into the spatial one. 
 
In order to integrate temporal information at different times, we can provide different frames with a unique interlacing offset. Instead of habitually assigning each channel with a separately learnable offset, we adopt distinctive offsets for different channel groups. As observed in SlowFast~\cite{feichtenhofer2018slowfast}, human perception on object motion focuses on different temporal resolutions. To maintain temporal fidelity and recognize spatial semantics jointly, different groups of temporal receptive fields pursuit a thorough separation of expertise convolution. Besides, groups of offsets also reduce the model complexity as well as stabilize the training procedure across heavy backbone architectures. 

\subsection{Deformable Shift Module}
\begin{figure}[htb]

\centering
\includegraphics[width=1.0\linewidth]{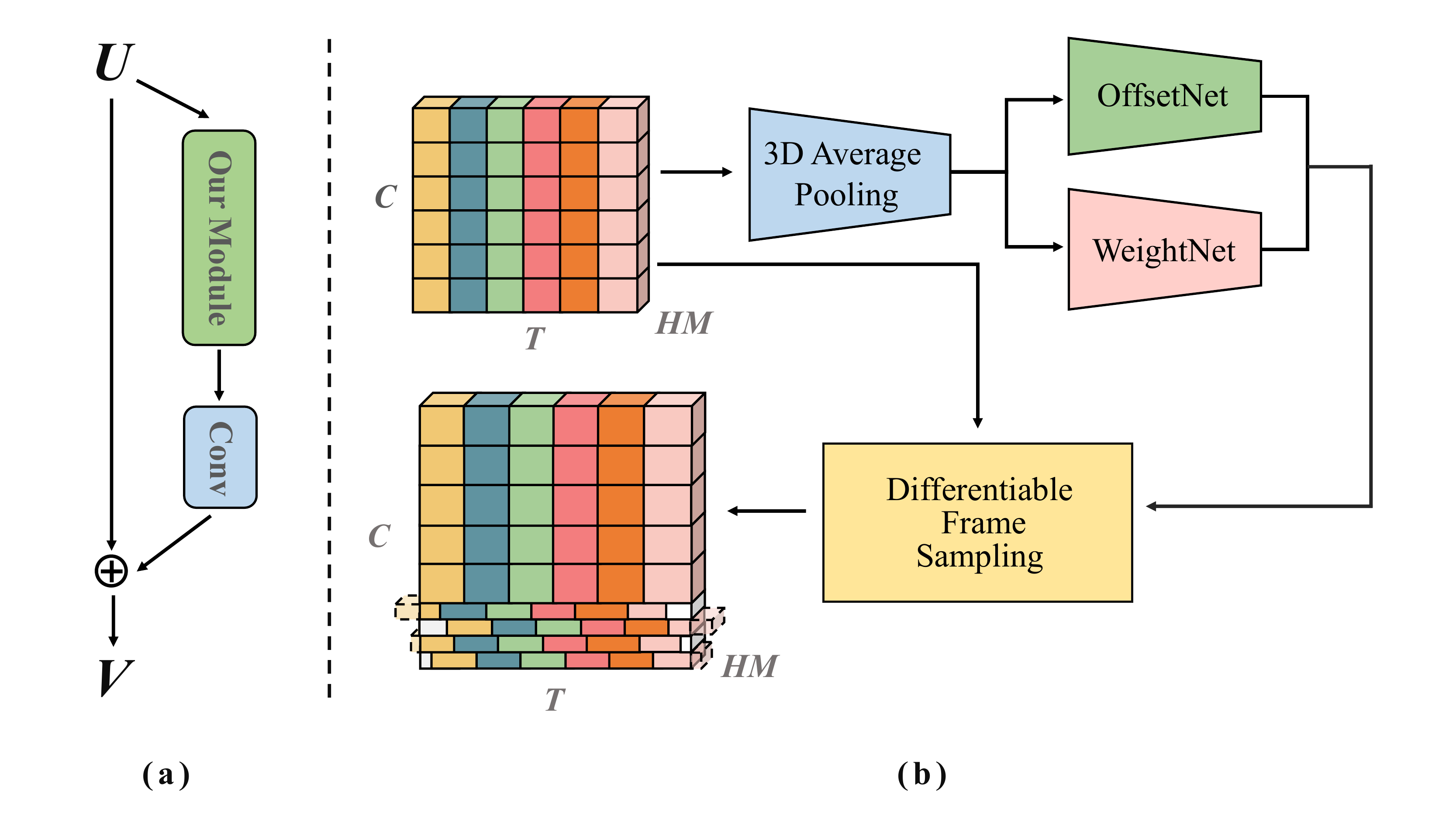}
\caption{Architecture of TIN: our input video clips composed of uniformly sampled from raw frames are processed by our modified 2D ResNet-50. \textbf{(a)} We plug our module before the convolution layer in each block. \textbf{(b)} Our module obtains offsets and weights by OffsetNet and WeightNet, then shifts and samples the feature along with the temporal dimension.}

\label{fig:archi}
\end{figure}
In our work, we fuse temporal information along the temporal dimension through inserting our module before each convolutional layer in the residual block~\cite{he2016deep}, as illustrated in Figure.~\ref{fig:archi}(a). Based on 2D CNN framework for video recognition, given a video, 8 or 16 frames are firstly uniformly sampled and then stacked as the input. TIN maps an input $U$ $\in \mathbb{R}^{T \times C \times H \times W}$ to the same size feature maps $V$ $\in \mathbb{R}^{T \times C \times H \times W}$ in each block. Here $N$ is the batch size, $T$ represents the temporal dimension. Besides, $C$ is the numbers of the channel, $H$ and $W$ refer to the spatial dimensions. In the notation that follows, we take $Offset_{g}$ to denote the learned offset and take $E_{g}$ to denote the learned weight, respectively, where $g$ means the parameters of the $g$-th group. Next, we will introduce our OffsetNet and WeightNet.


\textbf{OffsetNet} As shown in Figure.~\ref{fig:archi}(b), we firstly squeeze global spatial information into a temporal channel descriptor. The descriptor consists of one global average pooling layer, of which the kernel size is $H \times W$. We then transpose the $T \times C$ output to the shape of $C \times T$. The obtained representation only retains information on temporal and channel dimensions, which we termed it as $z$. The $c$,~$t$-th element of $z$ is computed by:

\begin{equation}\label{...}
z_{c,t} = Pooling(U_{c, t}) = \frac{1}{H \times W} \sum_{m=1}^{H}\sum_{n=1}^{W}U_{c,t}(m, n)
\end{equation}

To obtain the requisite parameters for the Differentiable Temporal Sampling Module, We exploit two paths to generate $Offset_{g}$ and $E_{g}$, respectively. The first path is to utilize the information extracted from pooling operation, taking a 1D convolution layer to aggregate the channel information. After the convolution operation, the feature map is transformed to $s$ $\in \mathbb{R}^{T}$. Then two fully-connected layers and ReLU~\cite{nair2010rectified} activation are applied to the output. The FC layers are capable of aggregating the information along the temporal dimension. The bias of the second fully-connected layer is initialized to make the post-sigmoid output start from 1.

\begin{equation}\label{2}
offset^{raw} = \sigma (W{2}\delta(W{1}(F_{1dconv}(z)))) \in \mathbb{R}^{G}
\end{equation}
In Equation~\ref{2}, $\delta$ denotes ReLU function,$F_{1Dconv}$ refers to the 1D convolution layer, $W_{1} \in \mathbb{R}^{T * T}$ , $W_{2} \in \mathbb{R}^{T * G}$ and G means the groups of channels. $\sigma$ here refers to the sigmoid function, which transforms the output to $(0, 1)$. In order to rescale the raw offset to the range of  $(-\frac{T}{2} , \frac{T}{2})$, scaling the output to this interval with sigmoid is found helpful for the stability and performance of the model. We rescale the offsets to this range for a global temporal receptive field.

\begin{equation}\label{3}
offset = (offset^{raw} - 0.5) \times T
\end{equation}

\textbf{WeightNet} Secondly, $E_{g}$ is calculated through the WeightNet. Our WeightNet consists of two parts: one convolution layer and a sigmoid function. The kernel size of the 1D convolution layer is 3, and the kernel numbers equal to the groups. Following the convolution layer, the sigmoid function and rescale module can scale our output to the range $(0, 2)$. Here we set the initial bias of the convolution layer to $0$, and the final initial output will be $1.0$.

\begin{figure}[htb]
\centering
\includegraphics[width=1.0\linewidth]{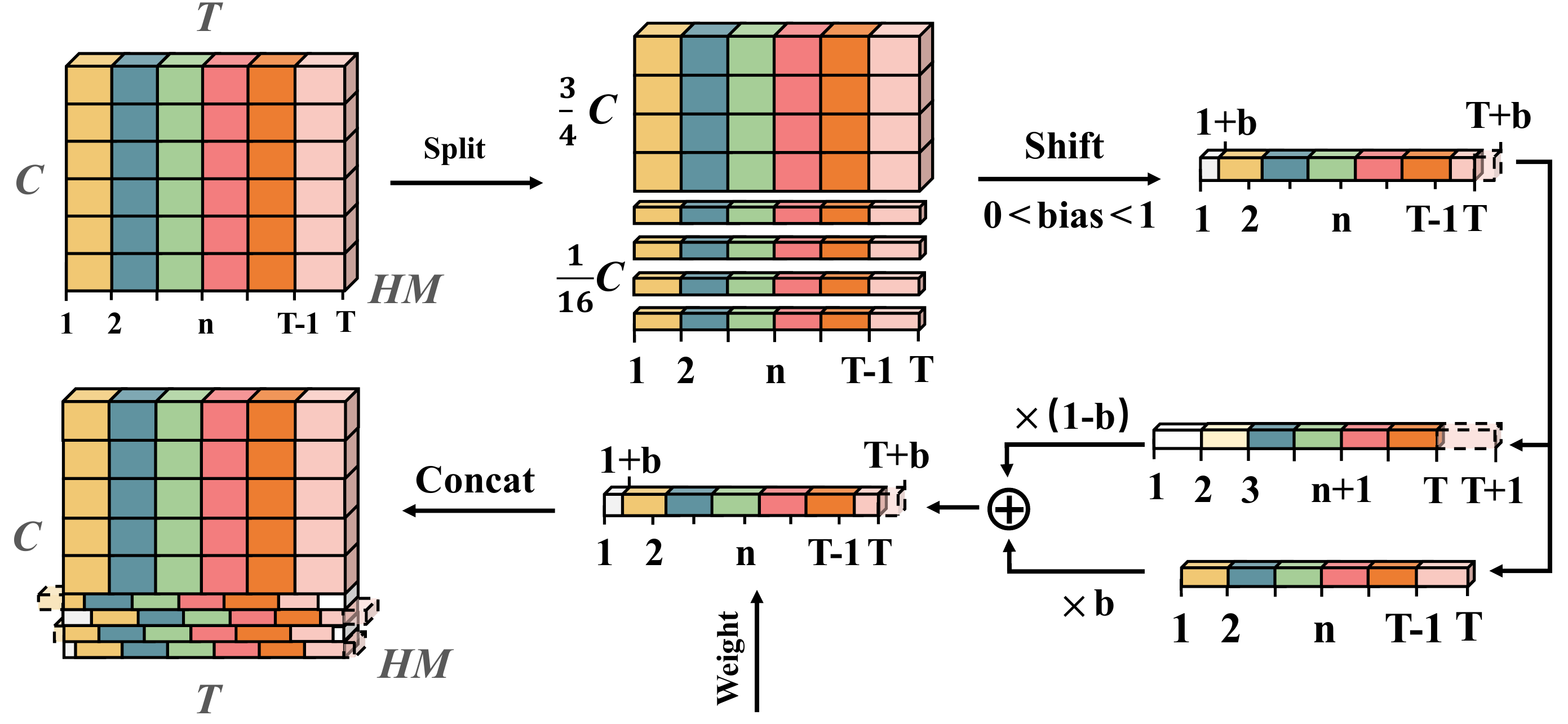}
\caption{The pipeline of differentiable Temporal-wise Frame Sampling. The module splits the feature map into several groups and shifts them by different offsets according to their group. Then compute the weighted sum along with the temporal dimension. Finally, we concatenate the split groups into the integral feature map, which is the same size as the input data.}
\label{fig:sample}
\end{figure}


\subsection{Differentiable Temporal-wise Frame Sampling}

To perform a shift along the temporal dimension of the input feature map, we design differentiable temporal-wise frame sampling. Above all, the input feature map $U$ is split into 2 parts along the channel dimension: one is to be shifted by different offsets according to different groupings, while the rest remains un-shifted.

As illustrated in Figure~\ref{fig:sample}, the shifted part is divided into four groups equally. In our work, we adopt a novel strategy to utilize symmetric offset. If we have $n$ groups, we only need to learn the offsets of half groups $\frac{n}{2}$, and the remained half are symmetrically derived by the previous offsets. In our setting, we keep $\frac{3}{4}C$ channels un-shifted and shift the rest $\frac{1}{4}C$ channels.

\textbf{Temporal-wise Frame Sampling} Notably, this step is essentially a linear interpolation process. For each group $g$ to be shifted, we assume that its offset $O_{g} \in (n_{0}, n_{0} + 1), n_{0} \in \mathbb{N}$, where $O_g$ denotes an arbitrary (fractional) values. The group of channels will be shifted two times, firstly we need to shift these channels by $n_{0}$ and re-shift them by $n_{0} + 1$. Then, we multiply the two sets based on the weights of $n_{0} + 1 - O_{g}$ and $O_{g} - n_{0}$ and add them together. This procedure is shown in Figure.~\ref{fig:archi}(b). Note that we assume the offset $\in (0, 1)$. At last, all shifted and un-shifted channels will be concatenated to feature map $V$ which shares the same size with input feature map $U$.

\begin{equation} \label{5}
\begin{split}
U_{c, t} = (n_{0} + &1 - O_{g})U_{c, t + n_{0}} + (O_{g} - n_{0})U_{c, t + n_{0} + 1} \\
&O_{g} \in (n_{0}, n_{0} + 1), n_{0} \in \mathbb{N}
\end{split}
\end{equation}

\textbf{Temporal Extension}
Meanwhile, some features may be shifted out and become ZERO due to the shift operator, further losing their gradient during training. Note that the temporal range of our input is $[1, T]$. To alleviate this problem, we set a buffer to help the shifted feature that falls in the $(0, 1)$ and $(T, T + 1)$ intervals. When the shifted time exceeds $T + 1$ or is less than 0, it will be clamped to 0(e.g. \ $U_{c, t=0.4} = 0.6 * U_{c, t=0} + 0.4 * U_{c, t=1.0} $ instead of 0). \\

\textbf{Temporal Attention}
As illustrated in Figure.~\ref{fig:archi}(b), when we concatenate the split groups of channels to $V$, the feature map is multiplied by the weight $E$ along the temporal dimension. Since some feature that belongs to both ends of the temporal dimension is possibly shifted out, attention mechanism is exploited to re-weight the feature and better capture long-range information.

As illustrated in Table.~\ref{groupnum}, we found TIN has the best performance when we divide the channels into four groups except for the un-shifted part, which consists of $\frac{3}{4}C$ channels. Among them, two groups use two offsets while the other two use their reversed offsets. 

\subsection{Analysis}
This work is built upon the idea of fusing a single frame with neighboring frames in the channel dimension. Temporal Fusion is completed through: 1.~Shifting groups of channels. 2.~Temporal attention with the offsets and weights learned from target tasks. To help the symmetric flow of information in the temporal dimension, we adopt the strategy of reverse offsets. This prior brings an acceleration of convergence and performance elevation.

\textbf{Regularized Temporal Convolution Network} \noindent In this section, we prove that TIN is theoretically equivalent to the Regularized Temporal Convolution Network. We assume that the number of groups is equal to channel numbers in the following proof. For the sake of understanding, we have adopted a similar proof as in DCN. The convolution operator can be composed of two steps: 1) using a regular slice S to sample the input feature map along the temporal dimension. 2)summing up the sampled values and multiply them with weight. The slice S defines the receptive filed shape. We firstly take a plain 1D convolution as an example, 

\begin{equation} 
S = \{(-1), (0), (1)\}
\end{equation}
defines a kernel with the size of 3. \\
For each element $t_{0}$ along the output temporal dimension of feature map $V$, we have the definition
\begin{equation} \label{15}
V_{t_{0}} = \sum_{s_{n} \in S}^{ } w(s_{n}) \times U_{s_{n} + t_{0}}
\end{equation}
where $s_{n}$ represents locations in S, $w(t_{0})$ denotes the learned weight on the temporal dimension. In TIN, the regular slice S will be accumulated with $O{c}$, where $c$ means the number of groups. The feature in the same channel share the same offset. Thus the formulation of TIN is defined as:  \\
\begin{equation} \label{14}
V_{t_{0}, c} = w(t_{0}) \times U_{t_{0} + O_{c}} 
\end{equation}
 
$t_{0}$ is not an integer and $O_{c} \in (n_{0}, n_{0} + 1), n_{0} \in \mathbb{N}$:
 \begin{equation} \label{8}
 \begin{split}
V_{t_{0}, c} = w(t_{0})&\times((O_{c}-n_{0}) \times U_{t_{n_{0}+1}, c}+ \\
&(n_{0}+1-O_{c}) \times U_{t_{n_{0}}, c})
\end{split}
\end{equation}

We can change Equation 4 into another form which is similar to the definition of 1D convolution :

\begin{equation} \label{11}
w_{c}^{'}= [(n_{0}+1-O_{c})\times w(t_{0}), (O_{c}-n_{0})\times w(t_{0})]
\end{equation}

\begin{equation} \label{12}
S_{c}^{'} = [n_{0}, n_{0} + 1], n_{0} \in \mathbb{N}
\end{equation}

\begin{equation} \label{13}
V_{t_{0}, c} = \sum_{s_{n} \in S_{c}^{'}}^{ } w_{c}^{'}(s_{n}) \times U_{s_{n} + t_{0}}
\end{equation}

From Eq.~\ref{13} and Eq.~\ref{15}, we can find that TIN is converted to a constrained convolution kernel with a kernel size of 2. Specifically, the divided groups of channels in our framework indicates that inter-group channel possesses different offset while the intra-group channels share the same offset. 

Intuitively, different strategies of grouping lead to distinctive numbers of equivalent convolution kernels, which further validates the equivalence of proposed TIN and r-TCN.

 Proposed arbitrary offsets also help the deformable convolution kernels have a global receptive field. And with TIN, the follow-up network can obtain features from adjacent frames at specific time by learning their respective offset and weight. For example, if the learned offset are $O_{1}$, $O_{2}$, $-O_{1}$, $-O_{2}$, and $O_{1} \in (n_{0}, n_{0} + 1)$, $O_{2} \in ({n_{1}, n{_1} + 1})$, then in one channel we have feature from $t_{0}, t_{0} + n_{0}, t_{0} + n_{0} + 1, t_{0} + n{_1}, t_{0} + n{_1} + 1$ (assume no overlapping exists). In this way, each channel mingles information from different time stamps, further improving feature quality and facilitating temporal modeling.
 
 We take ResNet-50 as our backbone. The next layer after our module is a 1$\times$1 2D convolution layer, which can be seen as a Fully-Connected layer if we only consider the temporal dimension, which integrates information from different times. With the dynamic temporal receptive fields proposed by TIN, it can further mix up the frames into short video segments that can help capture more precise temporal information. In contrast, it can only obtain information from the fixed timetable without our design.\\
 
 \section{Experiments}
In this section, we demonstrate the effectiveness of the proposed TIN on many video datasets. We first introduce the datasets used in our experiments. Then we provide a quantitative analysis with 2D CNN baseline and TSM~\cite{lin2018temporal}. We also perform comparisons with the sota results on the dataset Something (V1 $\&$ V2). To conclude, we conduct ablation experiments about our TIN and study the functionality of our design.

\subsection{Setup and Implementation Details}
\textbf{Datasets} we conduct experiments on six video recognition datasets, including Something-Something (V1 $\&$ V2)~\cite{goyal2017something}, Kinetics-600~\cite{carreira2017quo}~\cite{carreira2018short}, UCF101~\cite{soomro2012ucf101}, HMDB51~\cite{carreira2017quo}, Multi-Moments in Time~\cite{monfort2019multimoments} and Jester datasets. Among them, K600 is a large action dataset that has 30k validation videos in 600 classes and ~392k training videos. Multi-Moments in Time is a large-scale and multi-label video dataset which includes over two million action labels for over one million three second videos. The Something and Jester datasets focus more on temporal modeling and the relationship inside video sequences. The labels of these videos are abstract like ``Dropping something in front of something'' and do not dependent on specific events such as ``air drumming'', ``yoga''. The labels of K600, UCF101, HMDB51, MMiT are more concrete.\\
\textbf{Details}
We take ResNet-50 as our backbone for the fair comparison with the state-of-the-art methods. Specifically, our model takes segments with a size of $T \times 224 \times 224$. 224 is the height and width of cropped frames. T is the number of frames that refers to 8 or 16 in our setting. To facilitate comparison with TSN, we do not downsample along the temporal dimension. Regarding data augmentation, we use spatial jittering and horizontal flipping to alleviate over-fitting. Due to many video dataset are not large enough, we set the dropout rate~\cite{srivastava2014dropout} to 0.5 and set weight decay to 5e-4. We use the mini-batch stochastic gradient descent~\cite{bottou2010large} algorithm with a momentum of 0.9 as our optimizer. We follow the common practice~\cite{wang2016temporal,wang2018non,qian2019aggregation} to initialize our network from pre-trained models on Kinetics~\cite{carreira2017quo} and freeze Batch Normalization~\cite{ioffe2015batch}. The initial learning rate is set to 0.005 and divided by 10 at 10, 20 epochs, which stops at 25 epochs. \\
We use the same inference settings in~\cite{lin2018temporal} which not only uses the same number of frames for both training and testing, but also takes only one center crop in each frame. \\

\begin{table*}[htb]
\begin{center}
\resizebox{0.8\textwidth}{!}{
\begin{tabular}{cccccccc}
\Xhline{1.2pt}
\textbf{Model}         & \textbf{Backbone} & \textbf{\#Frame} & \textbf{FLOPs/Video} & \textbf{\#Param.} & \textbf{Val Top-1} & \textbf{Val Top-5}  \\ \Xhline{1.2pt}
TSN                    & BNInception       & 8                & 16G                  & 10.7M             & 19.5               &   -                                      \\
TSN                    & ResNet-50         & 8                & 33G                  & 24.3M             & 19.7               & 46.6                                    \\
TRN-Multiscale         & BNInception       & 8                & 16G                  & 18.3M             & 34.4               &   -                               \\
TRN-Multiscale         & ResNet-50         & 8                & 33G                  & 31.8M             & 38.9               & 68.1                                 \\
Two-stream TRN$_{RGB+Flow}$ & BNInception       & 8+8              & -                    & 36.6M             & 42.0               &                                \\ \hline
ECO                    & BNIncep+3D Res18  & 8                & 32G                  & 47.5M             & 39.6               &   -                                      \\
ECO                    & BNIncep+3D Res18  & 16               & 64G                  & 47.5M             & 41.4               &   -                                    \\
ECOEnLite              & BNIncep+3D Res18  & 92               & 267G                 & 150M              & 46.4               &   -                               \\
ECOEnLite$_{RGB+Flow}$      & BNIncep+3D Res18  & 92+92            & -                    & 300M              & 49.5               &            -                   \\ \hline
I3D                    & 3D ResNet-50      & 32*2clip         & 153G * 2             & 28.0M             & 41.6               & 72.2                                   \\
Non-local I3D          & 3D ResNet-50      & 32*2clip         & 168G * 2             & 35.3M             & 44.4               & 76.0                                    \\
Non-local I3D + GCN    & 3D ResNet-50+GCN  & 32*2clip         & 303G * 2             & 62.2M             & 46.1               & 76.8                               \\ \hline
TSM                    & ResNet-50         & 8                & 33G                  & 24.3M             & 43.4               & 73.2                                   \\
TSM                    & ResNet-50         & 16               & 65G                  & 24.3M             & 44.8               & 74.5                                   \\
TSM$_{En}$                   & ResNet-50         & 24               & 98G                  & 48.6M             & 46.8               & 76.1                                    \\
            \hline
            
TIN                   & ResNet-50         & 8                & 34G                  & 24.3M             & \textbf{45.8}      &  \textbf{75.1}                                      \\
TIN                   & ResNet-50         & 16               & 67G                  & 24.3M             & \textbf{47.0}      &   \textbf{76.5}                                      \\
TIN$_{En}$                 & ResNet-50         & 24               & 101G                 & 48.6M             & \textbf{49.6}     & \textbf{78.3}                     \\
                                 \Xhline{1.2pt}
                       &                   &                  &                      &                   &                    &                    &                    
\end{tabular}
}
\end{center}
\caption{Quantitative comparison of TIN with other methods on Something-Something v1 dataset.}
\label{something}
\end{table*}

\subsection{Comparison with 2D CNN baseline and TSM}
The Something-Something (V1) dataset shows the abstract pre-defined action which human perform with common objects. The dataset consists of ~108k videos of 174 classes. It focuses on the modeling of Time Series. Traditional 2D models have inferior performances on this dataset since their basic architectures cannot capture complex temporal interactions and get confused with the category in reverse order(e.g. ``Pushing something from left to right'' and ``Pushing something from right to lest''). \\
In this section, we demonstrate that TIN can help TSN improve significantly with little more computation and parameters overhead. We conduct comparisons with TSM in Table.~\ref{2dbase}, Table.~\ref{mmit} and Table.~\ref{something}. In Table.~\ref{2dbase}, we report the results on three temporal-sensitive datasets: Something-Something V1, Something-Something V2 and Jester. On Something V1, TIN achieves 26.6\% better performance compared to 2D TSN baseline and 1.5\% better performance than TSM. In Something V2 and Jester,  TIN also outperforms other approaches by a large margin. \\
We also show the results on four temporal-insensitive datasets: Kinetics-600, HMDB51, UCF101, Multi-Moments in Time. TIN consistently outperforms the 2D baseline and stronger TSM. For instance, on the large-scale dataset Kinetics-600, TIN obtains 0.9\% better performance than TSM and 1.7\% better performance than 2D baseline. TIN is also one of the main modules of our top-1 solution in ICCV19 Multi-Moments in Time (MIT) Challenge. In order to demonstrate the efficiency of TIN, We also report the accuracy, the parameters and FLOPs in Table.~\ref{something}.

\begin{table}[htb]
\begin{center}
\resizebox{0.7\columnwidth}{!}{
\begin{tabular}{lllll}
\Xhline{1.2pt}
\textbf{Dataset}              & \textbf{Model} & \textbf{Acc1} & \textbf{Acc5}  \\ \Xhline{1.2pt}
\multirow{3}{*}{Something V1  ~~~~~~~~~~~~~ } & TSN~~~~   & 20.4~~     & 48.1~~                               \\
                              & TSM   &  45.5   & 75.4                               \\
                              & Ours  & \textbf{47.0}    & \textbf{96.6}                               \\ \hline
\multirow{3}{*}{Something V2} & TSN   & 32.0     &  63.2                           \\
                              & TSM   & 59.2     &  85.3                              \\
                              & Ours  & \textbf{60.0}     &  \textbf{85.5}                             \\ \hline
\multirow{3}{*}{Jester}       & TSN   & 82.3     &  99.2                            \\
                              & TSM   & 96.2    &  99.7                               \\
                              & Ours  &  \textbf{96.7}    & \textbf{99.8}                              \\ \hline
\multirow{3}{*}{UCF101}       & TSN   &  90.1    &  98.7                           \\
                              & TSM   &  92.6    & 99.1                            \\
                              & Ours  &  \textbf{93.6}    & \textbf{99.1}                             \\ \hline
\multirow{3}{*}{HMDB51}       & TSN   &   63.9   & 88.2                                 \\
                              & TSM   &  69.5    &  90.3                                \\
                              & Ours  &  \textbf{72.0}    &  \textbf{92.0}                             \\ \hline
\multirow{3}{*}{Kinetics-600} & TSN   &  68.7    &  88.2                                \\
                              & TSM   & 69.5     &  89.5                               \\
                              & Ours  &  \textbf{70.4}    &  \textbf{89.6}                           \\ \Xhline{1.2pt}
\end{tabular}
}
\end{center}

\caption{A comparison between TIN, TSM and 2D baseline TSN. It shows TIN has better performance than the others. All the below experiments were performed under the same training settings with ResNet-50 backbone.} 

\label{2dbase}
\end{table}

\begin{table}[]
\begin{center}
\resizebox{0.7\columnwidth}{!}{
\begin{tabular}{c|llc}
\Xhline{1.2pt}
\textbf{Method} & \textbf{Backbone}   & \textbf{\#Frame} & \textbf{mAP}   \\ \Xhline{1.2pt}
TSN    & ResNet-101 & 5       & 58.92 \\
TSM    & ResNet-101 & 8       & 61.06 \\
TIN    & ResNet-101 & 8       & \textbf{62.22} \\ \Xhline{1.2pt}
\end{tabular}
}
\end{center}
\caption{mAP on the validation set of the Multi Moments in Time Dataset.}
\label{mmit}
\end{table}

\subsection{Comparison with State-of-the-arts}
 \begin{figure}[htb]
\centering
\includegraphics[width=0.8\linewidth]{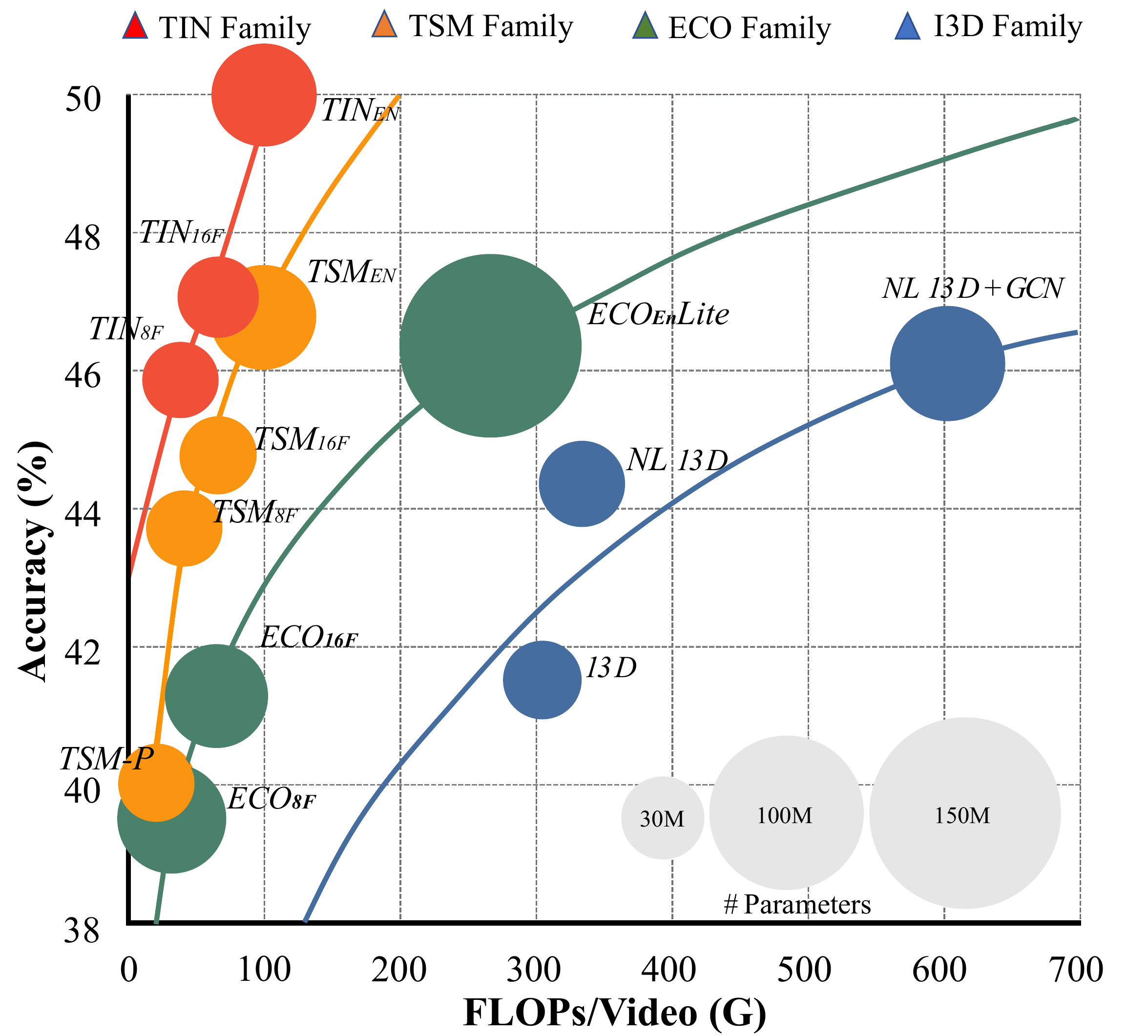}

\caption{Comparison about the trade-offs between the accuracy, the parameters and FLOPs}

\label{fig:demo}
\end{figure}

In this section, we show that TIN does not only have better performance than 2D baseline TSN and Temporal Shift Network, but also produces the best performance compared with state-of-the-art methods on the datasets that focusing on temporal modeling.

\textbf{Evaluation on Something-Something v1} We conduct comparison with Temporal Segment Network (TSN)~\cite{wang2016temporal}, Temporal Relation Networks (TRN)~\cite{zhou2018temporal}, Efficient Convolution Network for Online Video Understanding(ECO)~\cite{zolfaghari2018eco}, Inflated 3D Network (I3D)~\cite{carreira2017quo}, Non-local Neural Networks~\cite{wang2018non} and Temporal Shift Module (TSM)~\cite{lin2018temporal}. Table.~\ref{something} reports the result of proposed TIN and other representative methods. Our TIN has an increasing inherence in modeling temporal information. We start with the setting that the frames are 8 and the backbone is ResNet-50. 

TSN produces unsatisfying results since it is not capable of modeling temporal sequences. TRN fuses temporal information at high level after  feature extraction and achieves 19.2\% better performance than TSN. The previous state-of-the-art TSM can fuse temporal information at all stages and brings 1.4\% improvement. Our proposed module TIN can shift adaptive offsets according to specific datasets as well as capture long-range information. Therefore it introduces 2.4\% improvement to the current state-of-the-art TSM. 

\begin{figure*}[htb]
\centering
\includegraphics[width=0.9\linewidth]{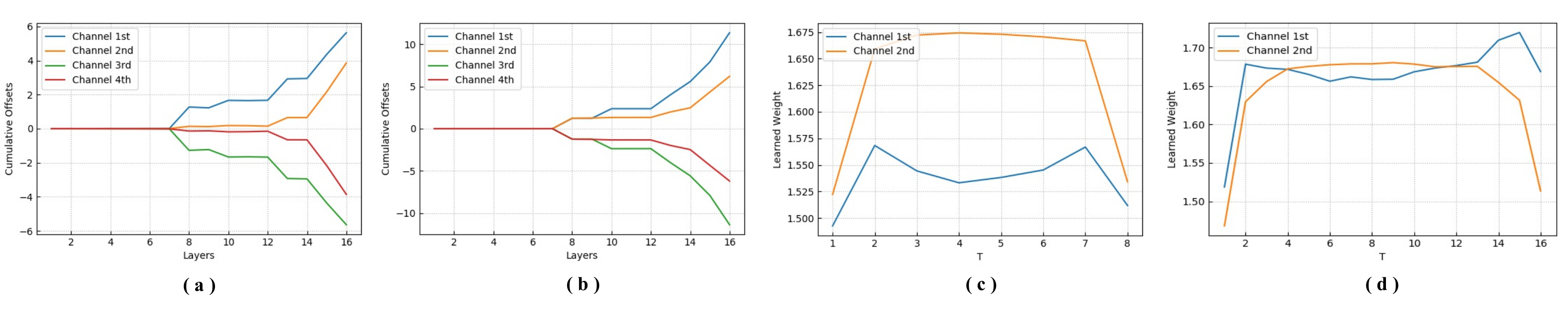}

\caption{Examples of the distribution of our learned offsets and weights from 8-frames input model and 16-frames model. For each group's offset, we plot lines to shows cumulative offsets on layers with different depths in (a)(b). We also present the learned weights in the temporal dimension in (c)(d).}

\label{fig:distribution}
\end{figure*}

ECO~\cite{zolfaghari2018eco} focuses on the efficiency in video understanding. It is proposed to merge long-term content rather than late temporal fusion. Its FLOPs is smaller than I3d, Non-local I3D, and other 3D-Based models. When the input consists of 16 frames, TIN still outperforms ECO significantly. ECO ensembles a model that is trained with \{16, 20, 24, 32\} number of frames. However, the parameters and FLOPs are too large for deployment. The ensembled model has 6$\times$ more parameters and 4$\times$ more FLOPs than our 16-frames model. In order to facilitate the comparison of results, we also provide one ensembled model trained from \{8, 16\} frames. It has a 3.6\% clear improvement compared to ECO with 2.6 $\times$ fewer FLOPs and 3 $\times$ fewer network parameters. \\

Non-local I3D + GCN is a previously state-of-the-art method. The method can capture long-range dependencies and process the temporal information at all stages. Compared with Non-local I3D, our 16-frames model gets 2.6\% better performance and 5 $\times$ fewer FLOPs. We suppose that the  Non-local module uses lots of parameters to model the feature in time and space dimension. However, Something dataset mainly focuses on temporal modeling and is not sensitive to spatial information.

Temporal Shift Module (TSM) is a strong baseline in this dataset. TIN also obtains comparable performance to it. Our 8-frames model and 16-frames model have a ~2\% improvement compared to TSM's. Particularly, ensembled TIN using \{8, 16\} frames has about 2.8\% better performance than TSM with almost the same FLOPs and parameters.

\begin{table}[]
\centering
\resizebox{0.7\columnwidth}{!}{
\begin{tabular}{lcc}
\Xhline{1.2pt}
\textbf{Method}       & \textbf{Val Top-1} & \textbf{Val Top-5} \\ \Xhline{1.2pt}
TSN                   & 30.0               & 60.5               \\ \hline
MultiScale TRN        & 48.8               & 77.6               \\
2-Stream TRN          & 55.5               & 83.1               \\ \hline
TSM$_{16f}$             & 59.4               & 86.1               \\ \hline
TIN$_{16f}$ & \textbf{60.1}      &  \textbf{86.4}      \\ \Xhline{1.2pt}        
\end{tabular}}
\caption{Comparison of TIN against other methods on Something-Something v2 dataset.}
\label{somethingv2}
\end{table}

\textbf{Evaluation on Something-Something v2} \noindent Something v2 is a larger dataset compared to v1, the number of videos has been greatly increased and the label noise has been greatly reduced. We report the respective results of TIN, TRN and TSM in Table.~\ref{somethingv2}, where TIN also obtains a state-of-the-art performance and is leaving a 0.7\% gap to previous State-of-the-arts.

\subsection{Visualization of learned offsets and weights}
In addition to video recognition accuracies, we want to attain further insight into the learned offsets. Note that we take Image-Net pre-trained model that will not influence the learned temporal offsets, to remove the impact of the kinetics pre-trained model. Our backbone is ResNet-50 and has 4 blocks, with \{4, 4, 4, 4\} layers in each block. We accumulate the mean offsets of each layer per group and provide the different behavior of models with different frame inputs. 

\textbf{Learned Offsets}  In Figure.~\ref{fig:distribution}(a)(b), the offsets in the first seven layers are almost zero, and we suppose that shallower layers mainly focus on learning the spatial information, since earlier temporal shifting may influence the ability of the network to process spatial information. As shown in Figure.~\ref{fig:distribution}, the offsets after the first 7 layers gradually increase. The result also indicates that temporal modeling on high-level semantic features can have better performance. \\
From the comparisons between Figure.~\ref{fig:distribution}(a)(b), we can find that when the input frames change from 16 to 8 frames, the learned offsets also increase accordingly. It demonstrates that our module learns that temporal information fusion on higher levels can bring more improvement.

\textbf{Learned Weights}  As shown in Figure.~\ref{fig:distribution}(c)(d), the weights of the first and the last frames are significantly lower than others'. This phenomenon shows that frames at both ends lose partial information as it is shifted out. The weight module of TIN adjusts the weights of frames at both ends to reduce the impact on network training.

\subsection{Ablation Experiments}
This section provides ablation studies on Something v1, which focuses more on temporal interactions. \\
\begin{table}[]
\centering
\resizebox{0.8\columnwidth}{!}{
\begin{tabular}{llcc}
\Xhline{1.2pt}
\textbf{Groups} & \textbf{Pretrain} & \multicolumn{1}{l}{\textbf{With Reverse Offsets}} & \multicolumn{1}{l}{\textbf{Val Top-1}} \\ \Xhline{1.2pt}
1               & ImageNet          & True                                             & 42.0                                   \\
2               & ImageNet          & True                                             & \textbf{42.9}                                   \\
2               & ImageNet          & False                                             & 41.1                                   \\
4               & ImageNet          & True                                             & 42.4                                   \\
4               & ImageNet          & False                                            & 42.6                                   \\ \Xhline{1.2pt}
\end{tabular}
}
\caption{Accuracies with different numbers of groups and reverse offsets}

\label{groupnum}
\end{table}

\textbf{Groups of shifted channels} In Table.~\ref{groupnum}, We firstly study the effect of group numbers. The best performance is obtained when the group number is 2. Note that the group numbers in Tale~\ref{groupnum} do not include the reverse groups. If the number increases or decreases, the performance will all drop. This phenomenon indicates that: if the number is too small, there will be fewer information from distinctive times stages when integrating temporal information, thus unable to model complex temporal dynamics; if the number goes too large, it introduces much more context to integrate as well as more brittle training process.  \\

\textbf{Reverse offsets} We assume that the reverse offsets can help interlacing temporal information symmetrically. It equip TIN with a symmetric receptive field. Results in Table.~\ref{groupnum} further verify our hypothesis. The performance of the second model with 2 groups and 2 reverse groups has 0.3\% improvement than the fourth model which has 4 groups and 0 reverse groups.

\section{Conclusion}

We propose a temporal interlacing network(TIN), a simple yet powerful operator in temporal modeling. The learnable shifting module learns to interlace temporal and spatial information jointly with little computation overhead. TIN endows learned feature with both accurate and efficient representation, advancing dominating results on video classification.

\bibliographystyle{aaai}
{\small \bibliography{citation}}
\end{document}